\documentclass[11pt]{article}
\usepackage{acl2015}
\usepackage{times}
\usepackage{url}
\usepackage{latexsym}
\usepackage{bm}
\usepackage{amsmath}
\usepackage{amsfonts}
\usepackage{graphicx}
\usepackage{delarray}
\usepackage{algorithm}
\usepackage{algpseudocode}
\usepackage{graphicx}
\usepackage{subfigure}
\usepackage{multirow}
\usepackage{float}

\title{Efficient Learning for Undirected Topic Models}

\author{ Jiatao Gu and Victor O.K. Li
         \\ Department of Electrical and Electronic Engineering
         \\ The University of Hong Kong
         \\  {\tt \{jiataogu, vli\}@eee.hku.hk}
         \\}

\date{}

\begin{document}
\maketitle
\begin{abstract}
Replicated Softmax model, a well-known undirected topic model, is
powerful in extracting semantic representations of documents.
Traditional learning strategies such as Contrastive Divergence are very inefficient. 
This paper provides a novel estimator to speed up the learning based on Noise Contrastive Estimate, extended for documents of variant lengths and weighted inputs. 
Experiments on two benchmarks show that the new estimator achieves great learning efficiency and high accuracy on document retrieval and classification.

\end{abstract}

\section{Introduction}
Topic models are powerful probabilistic  graphical approaches to analyze document semantics in different
applications such as document categorization and information retrieval. They are mainly constructed by directed structure like pLSA \cite{hofmann2000learning} and LDA \cite{blei2003latent}. 
Accompanied by the vast developments in deep learning, several undirected topic models, such as \cite{salakhutdinov2009semantic,srivastava2013modeling}, have recently been reported to achieve great improvements in efficiency and accuracy. 

Replicated Softmax model (RSM)~\cite{hinton2009replicated}, 
a kind of typical undirected topic model, is
composed of a family of
Restricted Boltzmann Machines (RBMs). Commonly, RSM is learned like standard RBMs using approximate methods like Contrastive Divergence (CD).
However, CD is not really designed for RSM. 
Different from RBMs with binary input, RSM adopts softmax units to represent words, resulting in great inefficiency with sampling inside CD, especially for a large vocabulary.
Yet, NLP systems usually require vocabulary sizes of tens to hundreds of thousands, thus seriously limiting its application. 

Dealing with the large vocabulary size of the inputs is a serious problem in deep-learning-based NLP systems.
\newcite{bengio2003neural} pointed this problem out when normalizing the softmax probability in the neural language model (NNLM), and \newcite{morin2005hierarchical} solved it based on a hierarchical binary tree.
A similar architecture was used in word representations like \cite{mnih2009scalable,mikolov2013efficient}. 
Directed tree structures cannot be applied to undirected models like RSM, but stochastic approaches can work well.
For instance, \newcite{dahl2012training} found that several Metropolis Hastings sampling (MH) approaches approximate the softmax distribution in CD well, although MH requires additional complexity in computation.
\newcite{hyvarinen2007some} proposed Ratio Matching (RM) to train unnormalized models, and \newcite{dauphin2013stochastic} added stochastic approaches in RM to accommodate high-dimensional inputs.
Recently, a new estimator Noise Contrastive Estimate (NCE) \cite{gutmann2010noise} is proposed for unnormalized models, and shows great efficiency in learning word representations such as in \cite{mnih2012fast,mikolov2013distributed}.

In this paper, we propose an efficient learning strategy for RSM named 
$\alpha$-NCE, applying NCE as the basic estimator. 
Different from most related efforts that use NCE for predicting single word, our method extends NCE to generate noise for documents in variant lengths. It also enables RSM to use weighted inputs to improve the modelling ability. As RSM is usually used as the first layer in many deeper undirected models like Deep Boltzmann Machines \cite{srivastava2013modeling}, $\alpha$-NCE can be readily extended to learn them efficiently. 


\section{Replicated Softmax Model}
RSM is a typical undirected topic model, which is based on bag-of-words (BoW) to represent documents.
In general, it consists of a series of RBMs, each of which contains variant softmax visible units but the same binary hidden units.

Suppose $K$ is the vocabulary size.
For a document with $D$ words, if the $i^{th}$ word in the document equals the $k^{th}$ word of the dictionary, a vector $ \bm{v}_i \in \{0, 1\}^K$ is assigned, only with the $k^{th}$ element $v_{ik}=1$. 
An RBM is formed by assigning a hidden state $\bm{h} \in \{0, 1\}^H$ to this document $\bm{V} = \{ \bm{v}_1, ..., \bm{v}_D \}$, where the energy function is:
\begin{equation}
\label{eq:Eng}
E_{\bm{\theta}}(\bm{V}, \bm{h}) = -\bm{h}^T\bm{W}\bm{\hat{v}} - \bm{b}^T\bm{\hat{v}} - D\cdot \bm{a}^T\bm{h}
\end{equation}
where $\bm{\theta} = \{\bm{W}, \bm{b}, \bm{a} \}$ are parameters shared by all the RBMs, and $\bm{\hat{v}} = \sum_{i=1}^D{\bm{v}_{i}}$ is commonly referred to as the word count vector of a document. 
The probability for the document $\bm{V}$ is given by:
\begin{eqnarray}
\label{eq:Prb}
\begin{split}
P_{\bm{\theta}}(\bm{V}) &= \frac{1}{Z_D}e^{-F_{\bm{\theta}}(\bm{V})}, Z_D = \sum\nolimits_{\bm{V}}e^{-F_{\bm{\theta}}(\bm{V})}\\
F_{\bm{\theta}}(\bm{V}) &= \log{\sum\nolimits_{\bm{h}}e^{-E_{\bm{\theta}}(\bm{V}, \bm{h})}} 
\end{split}
\end{eqnarray}
where $F_{\bm{\theta}}(\bm{V})$ is the ``free energy", which can be analytically integrated easily, and $Z_D$ is the ``partition function" for normalization, only associated with the document length $D$.
As the hidden state and document are conditionally independent, the conditional distributions are derived:
\begin{align}
&P_{\bm{\theta}}\left(v_{ik}=1|\bm{h}\right) = 
\frac{\exp \left(\bm{W}_{k}^T\bm{h}+b_k \right)}
     {\sum_{k=1}^K 
      \exp \left(\bm{W}_{k}^T\bm{h}+b_k \right)}\\
&P_{\bm{\theta}}\left(h_j=1|\bm{V}\right) = 
\sigma\left(\bm{W}_{j}\bm{\hat{v}} + D\cdot a_j\right)
\end{align} 
where $\sigma(x) = \frac{1}{1+e^{-x}}$. 
Equation (3) is the softmax units describing the multinomial distribution of the words, and Equation (4) serves as an efficient inference from words to semantic meanings, where we adopt the probabilities of each hidden unit ``activated" as the topic features.

\subsection{Learning Strategies for RSM}
\label{CD}
RSM is naturally learned by minimizing the negative log-likelihood function (ML) as follows:
\begin{equation}
L(\bm{\theta}) = 
-\mathbb{E}_{\bm{V} \sim P_{\bm{data}}} 
\left[
\log P_{\bm{\theta}}(\bm{V})
\right]
\end{equation}
However, the gradient 
is intractable for the combinatorial normalization term $Z_D$.
Common strategies to overcome this intractability are MCMC-based approaches such as Contrastive Divergence (CD) \cite{hinton2002training} and Persistent CD (PCD) \cite{tieleman2008training}, both of which require repeating Gibbs steps of $\bm{h}^{(i)} \sim P_{\bm{\theta}}(\bm{h}|\bm{V}^{(i)})$ and $\bm{V}^{(i+1)} \sim P_{\bm{\theta}}(\bm{V}|\bm{h}^{(i)})$ to generate model samples to approximate the gradient. 
Typically, the performance and consistency improve when more steps are adopted.
Notwithstanding, even one Gibbs step is time consuming for RSM, since the multinomial sampling normally requires linear time computations. 
The ``alias method" \cite{kronmal1979alias} speeds up multinomial sampling to constant time while linear time is required for processing the distribution.
Since $P_{\bm{\theta}}(\bm{V}|\bm{h})$ changes at every iteration in CD, such methods cannot be used.

\section{Efficient Learning for RSM}
Unlike \cite{dahl2012training} that retains CD, we adopted NCE as the basic learning strategy. 
Considering RSM is designed for documents, we further modified NCE with two novel heuristics, developing the approach
``Partial Noise Uniform Contrastive Estimate" (or $\alpha$-NCE for short).

\subsection{Noise Contrastive Estimate}
Noise Contrastive Estimate (NCE), similar to CD, is another estimator for training models with intractable partition functions. 
NCE solves the intractability through treating the partition function $Z_D$ as an additional parameter $Z_D^c$ added to $\bm{\theta}$, which makes the likelihood
computable.
Yet, the model cannot be trained through ML as the likelihood tends to be arbitrarily large by setting $Z_D^c$ to huge numbers. Instead,
NCE learns the model in a proxy classification problem with noise samples.

Given a document collection (data) $\{ \bm{V}_d \}_{T_d}$, and another collection (noise) $\{ \bm{V}_n \}_{T_n}$ with $T_n = kT_d$, NCE distinguishes these $(1+k)T_d$ documents simply based on Bayes' Theorem, where we assumed data samples matched by our model, indicating $P_{\bm{\theta}} \simeq P_{\bm{data}}$, and noise samples generated from an artificial distribution $P_n$. 
Parameters are learned by minimizing the cross-entropy function:
\begin{eqnarray}
\begin{split}
J(\bm{\theta}) =-
&\mathbb{E}_{\bm{V}_d \sim P_{\bm{\theta}}} 
\left[
\log
\sigma_k(X(\bm{V}_d)) 
\right]
\\-
k&\mathbb{E}_{\bm{V}_n \sim P_n} 
\left[
\log		
\sigma_{k^{-1}}(-X(\bm{V}_n)) 
\right] 
\end{split}
\end{eqnarray}
and the gradient is derived as follows,
\begin{eqnarray}
\begin{split}
-\nabla_{\bm{\theta}}J(\bm{\theta}) = 
&\mathbb{E}_{\bm{V}_d \sim P_{\bm{\theta}}} 
\left[
\sigma_{k^{-1}}(-X) 
\nabla_{\bm{\theta}}X(\bm{V}_d)
\right] 
\\-
k&\mathbb{E}_{\bm{V}_n \sim P_n}  
\left[
	\sigma_k(X) 
\nabla_{\bm{\theta}}X(\bm{V}_n)
\right] 
\end{split}
\end{eqnarray}
where $\sigma_k(x) = \frac{1}{1+ke^{-x}}$, and the ``log-ratio" is:
\begin{equation}
X(\bm{V})=\log \left[ P_{\bm{\theta}}(\bm{V}) / P_n(\bm{V}) \right]
\end{equation}
$J(\bm{\theta})$ can be optimized efficiently with stochastic gradient descent (SGD). \newcite{gutmann2010noise} showed that the NCE gradient 
$\nabla_{\bm{\theta}}J(\bm{\theta})$ 
will reach the ML gradient when $k\rightarrow\infty$. In practice, a larger $k$ tends to train the model better.

\subsection{Partial Noise Sampling}
Different from \cite{mnih2012fast}, which generates noise per word, RSM requires the estimator to sample the noise at the document level.
An intuitive approach is to sample from the empirical distribution $\bm{\tilde{p}}$ for $D$ times, where the log probability is computed:
$\log P_n(\bm{V}) = \sum\nolimits_{\bm{v} \in \bm{V}} \left[\bm{v}^T\log\bm{\tilde{p}} \right]$.

For a fixed $k$, \newcite{gutmann2010noise} suggested choosing the noise close to the data for a sufficient learning result, indicating full noise might not be satisfactory. 
We proposed an alternative ``Partial Noise Sampling (PNS)" to generate noise by replacing part of the data with sampled words. 
\begin{algorithm}[h]
\caption{Partial Noise Sampling}
\begin{algorithmic}[1]
\State Initialize: $k, \alpha \in (0, 1)$
\For{each $\bm{V}_d = \{\bm{v}\}_{D} \in \{\bm{V}_d\}_{T_d}$}
\State Set: $D_r = \lceil \alpha \cdot D\rceil$
\State Draw: $\bm{V}_r = \{\bm{v}_r\}_{D_r} \subseteq \bm{V}$ uniformly
\For{$j = 1,...,k$}
\State Draw: $\bm{V}_n^{(j)} = \{\bm{v}_n^{(j)}\}_{D - D_r} \sim \bm{\tilde{p}}$
\State $\bm{V}_n^{(j)} = \bm{V}_n^{(j)} \cup \bm{V}_r$
\EndFor
\State Bind: $(\bm{V}_d, \bm{V}_r), (\bm{V}_n^{(1)}, \bm{V}_r),...,(\bm{V}_n^{(k)},\bm{V}_r)$
\EndFor
\end{algorithmic}
\end{algorithm}
See Algorithm 1, where we fixed the proportion of remaining words at $\alpha$, named ``noise level" of PNS.
However, 
traversing all the conditions to guess the remaining words 
requires $O(D!)$ computations.
To avoid this, we simply bound the remaining words with the data and noise in advance and the noise $\log P_{n}(\bm{V})$ is derived readily:\begin{equation}
\log P_{\bm{\theta}}(\bm{V}_r) +
\sum\nolimits_{\bm{v} \in \bm{V} \setminus \bm{V}_r} \left[\bm{v}^T\log\bm{\tilde{p}}\right] 
\end{equation}
where the remaining words $\bm{V}_r$ are still assumed to be described by RSM with a smaller document length. In this way, it also strengthens the robustness of RSM towards incomplete data.

Sampling the noise normally requires additional computational load.
Fortunately, since $\bm{\tilde{p}}$ is fixed, sampling is efficient using the ``alias method". It also allows storing the noise for subsequent use,
yielding much faster computation than CD.

\subsection{Uniform Contrastive Estimate}
When we initially implemented NCE for RSM, we found the document lengths terribly biased the log-ratio, resulting in bad parameters. Therefore
``Uniform Contrastive Estimate (UCE)" was proposed to accommodate variant document lengths by adding the uniform assumption:
\begin{equation}
\bar{X}(\bm{V}) = D^{-1}
\log \left[P_{\bm{\theta}}(\bm{V})/P_n(\bm{V}) \right]
\end{equation}
where UCE adopts the uniform probabilities $\sqrt[D]{P_{\bm{\theta}}}$ and $\sqrt[D]{P_n}$ for classification to average the modelling ability at word-level.
Note that 
$D$ is not necessarily an integer in UCE, and allows choosing a real-valued weights on the document such as \textit{idf}-weighting \cite{salton1983introduction}. Typically, it is defined as a weighting vector $\bm{w}$, where $w_k = \log \frac{T_d}{|\bm{V}\in\{\bm{V}_d\}: v_{ik}=1,\bm{v}_i\in \bm{V}|}$ is multiplied to the $k^{th}$ word in the dictionary. Thus for a weighted input $\bm{V}^w$ and corresponding length $D^w$, we derive: 
\begin{equation}
\tilde{X}(\bm{V}^w) = {D^w}^{-1} 
\log \left[P_{\bm{\theta}}(\bm{V}^w)/P_n(\bm{V}^w) \right]
\end{equation}
where
$\log P_n(\bm{V}^w) = \sum\nolimits_{\bm{v}^w \in \bm{V}^w} \left[ {\bm{v}^w}^T\log\bm{\tilde{p}} \right]$.
A specific $Z_{D^w}^c$ will be assigned to $P_{\bm{\theta}}(\bm{V}^w)$.

Combining PNS and UCE yields a new estimator for RSM, which we simply call $\alpha$-NCE\footnote{$\alpha$ comes from the noise level in PNS, but UCE is also the vital part of this estimator, which is absorbed in $\alpha$-NCE.}.
\section{Experiments}
\subsection{Datasets and Details of Learning}
We evaluated the new estimator to train RSMs on two text datasets: 20 Newsgroups and IMDB. 

The 20 Newsgroups\footnote{Available at http://qwone.com/\~{}jason/20Newsgroups} dataset is a collection of the Usenet posts, which contains 11,345 training and 7,531 testing instances. Both the training and testing sets are labeled into 20 classes. Removing stop words as well as stemming were performed.

The IMDB dataset\footnote{Available at http://ai.stanford.edu/\~{}amaas/data/sentiment} is a benchmark for sentiment analysis, which consists of 100,000 movie reviews taken from IMDB. The dataset is divided into 75,000 training instances ($1/3$ labeled and $2/3$ unlabeled) and 25,000 testing instances. Two types of labels, positive and negative, are given to show sentiment. Following \cite{maas2011learning}, no stop words are removed from this dataset. 

For each dataset, we randomly selected $10\%$ of the training set for validation, and the $idf$-weight vector is computed in advance.
In addition, replacing the word count $\hat{\bm{v}}$ by $\lceil \log\left(1+\hat{\bm{v}}\right)\rceil$ slightly improved the modelling performance for all models. 

We implemented 
$\alpha$-NCE
according to the parameter settings in \cite{hinton2010practical} using SGD in minibatches of size $128$ and an initialized learning rate of $0.1$.
The number of hidden units was fixed at $128$ for all models.
Although learning the partition function $Z_D^c$ separately for every length $D$ is nearly impossible, as in \cite{mnih2012fast} we also surprisingly found freezing $Z_D^c$ as a constant function of $D$ without updating never harmed but actually enhanced the performance.
It is probably because the large number of free parameters in RSM are forced to learn better when $Z_D^c$ is a constant. In practise, we set this constant function as
$Z_D^c = 2^H\cdot\left(\sum\nolimits_k{e^{b_k}}\right)^D$.
It can readily extend to learn RSM for real-valued weighted length $D^w$.

We also implemented CD with the same settings. All the experiments were run on a single GPU GTX970 using the library \textit{Theano} \cite{bergstra+al:2010-scipy}. To make the comparison fair, both $\alpha$-NCE and CD share the same implementation.



\subsection{Evaluation of Efficiency}
\label{learning}
To evaluate the efficiency in learning, we used the most frequent words as dictionaries with sizes ranging from $100$ to $20,000$ for both datasets, and test the computation time both for CD of variant Gibbs steps and 
$\alpha$-NCE
of variant noise sample sizes.
\begin{figure}[htbp]
\centering
\includegraphics[width=0.5\textwidth]{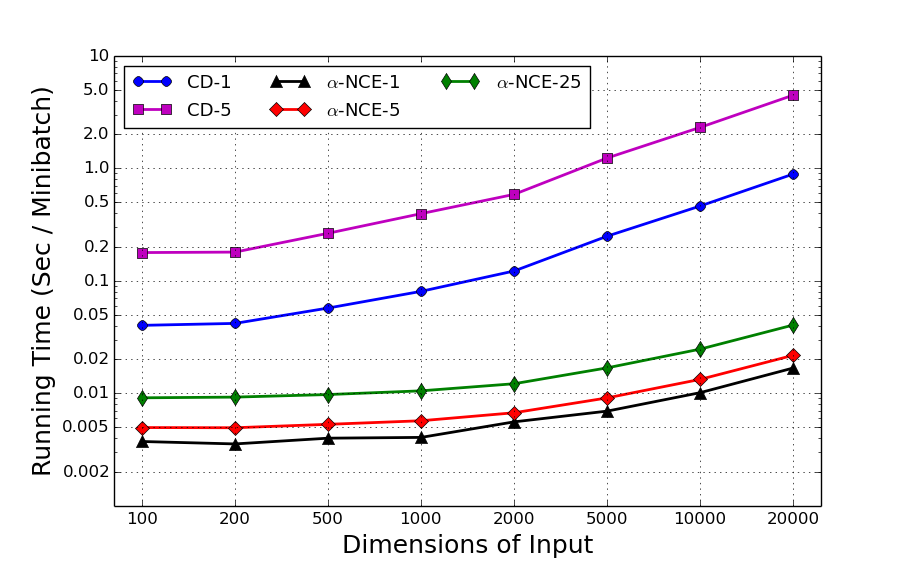}
\caption{\label{speedup}Comparison of running time}
\end{figure}
The comparison of the mean running time per minibatch is clearly shown in Figure \ref{speedup}, which is averaged on both datasets.
Typically,
$\alpha$-NCE
achieves $10$ to $500$ times speed-up compared to CD.
Although both CD and 
$\alpha$-NCE
run slower when the input dimension increases, CD tends to take much more time due to the multinomial sampling at each iteration, especially when more Gibbs steps are used. In contrast, running time stays reasonable in 
$\alpha$-NCE
even if a larger noise size or a larger dimension is applied.
 
\subsection{Evaluation of Performance}
One direct measure to evaluate the modelling performance is to assess RSM as a generative model to estimate 
the log-probability per word as \textit{perplexity}.
However, as 
$\alpha$-NCE
learns RSM by distinguishing the data and noise from their respective features, parameters are trained more like a feature extractor than a generative model. 
It is not fair to use \textit{perplexity} to evaluate the performance.
For this reason, we evaluated the modelling performance with some indirect measures.
\begin{figure}[htbp]
\centering
\includegraphics[width=0.45\textwidth]{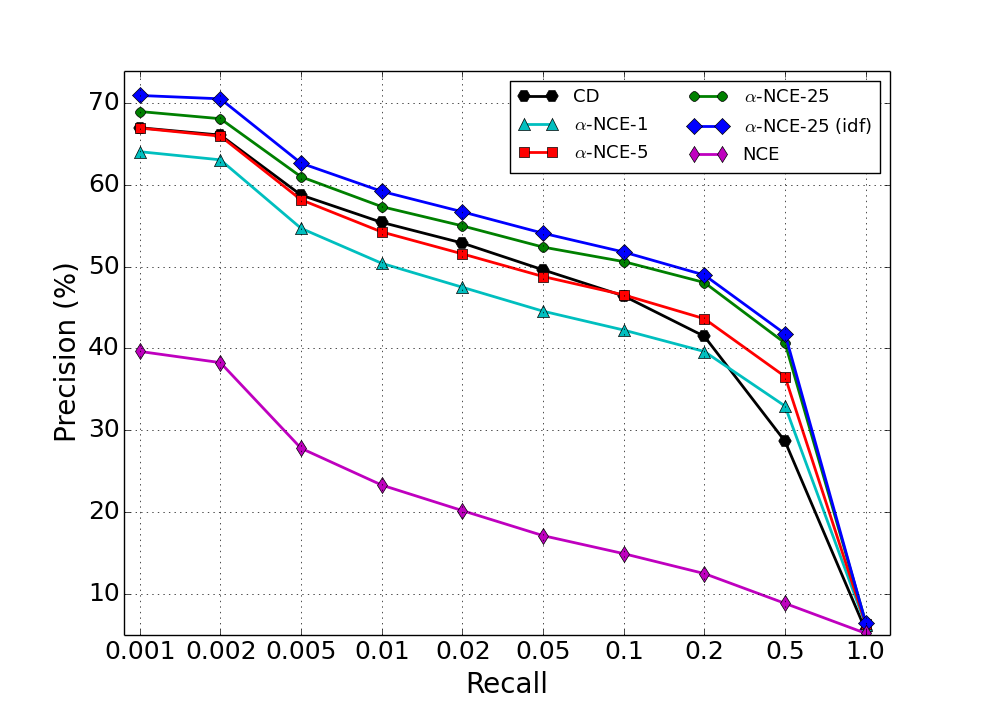}
\caption{\label{pr}Precision-Recall curves for the retrieval task on the 20 Newsgroups dataset using RSMs.}
\end{figure}

\begin{figure*}[htbp]
\centering
\subfigure[MAP for document retrieval]{
\label{pr2}
\includegraphics[width=0.315\textwidth]{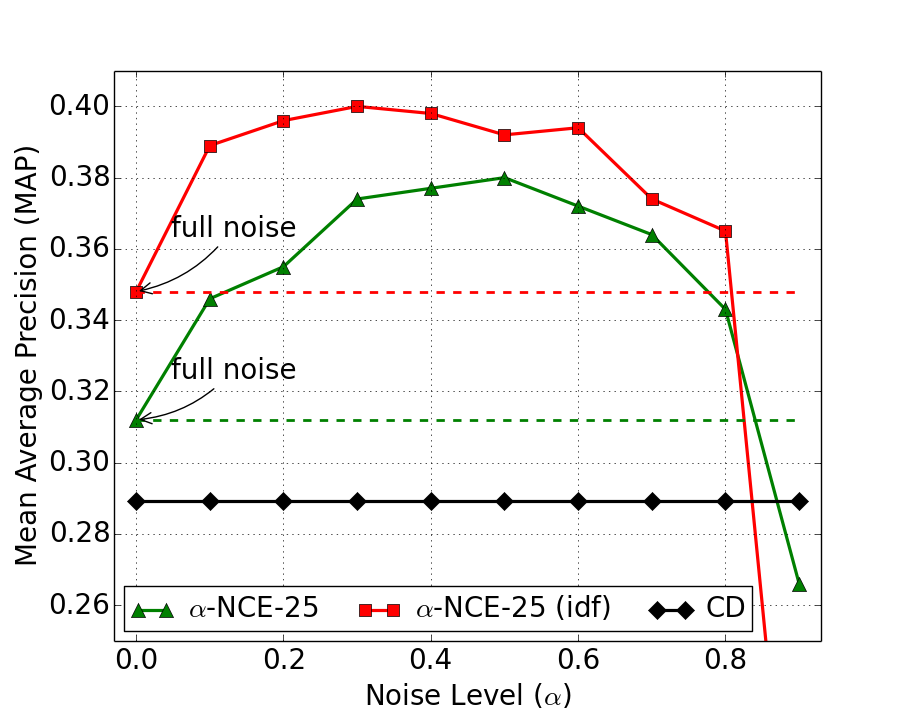}
}
\subfigure[Document classification accuracy]{
\label{pn}
\includegraphics[width=0.315\textwidth]{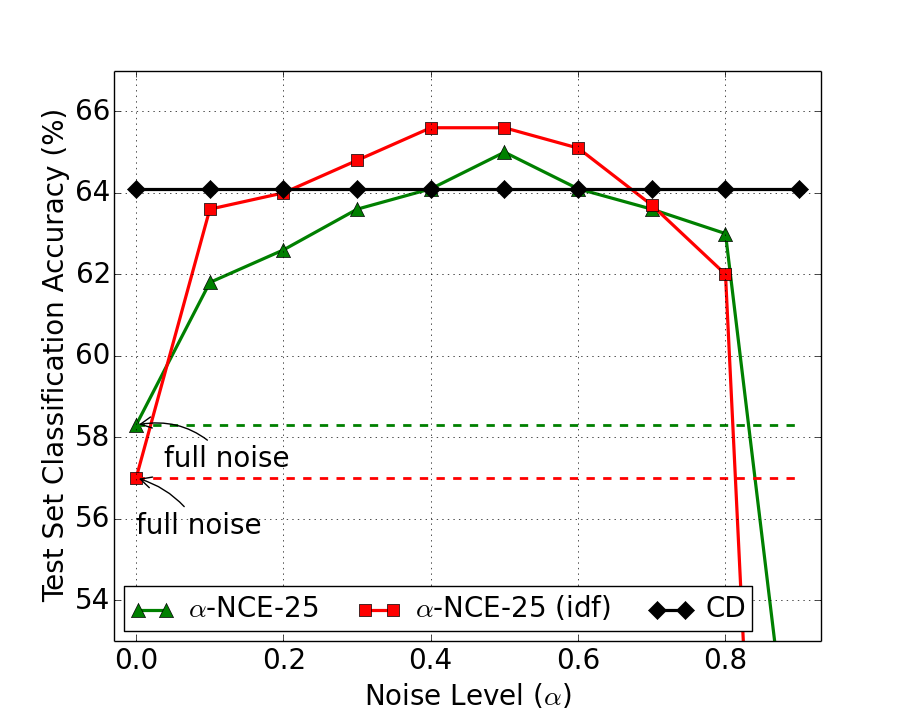}
}
\subfigure[Sentiment classification accuracy]{
\label{pc}
\includegraphics[width=0.315\textwidth]{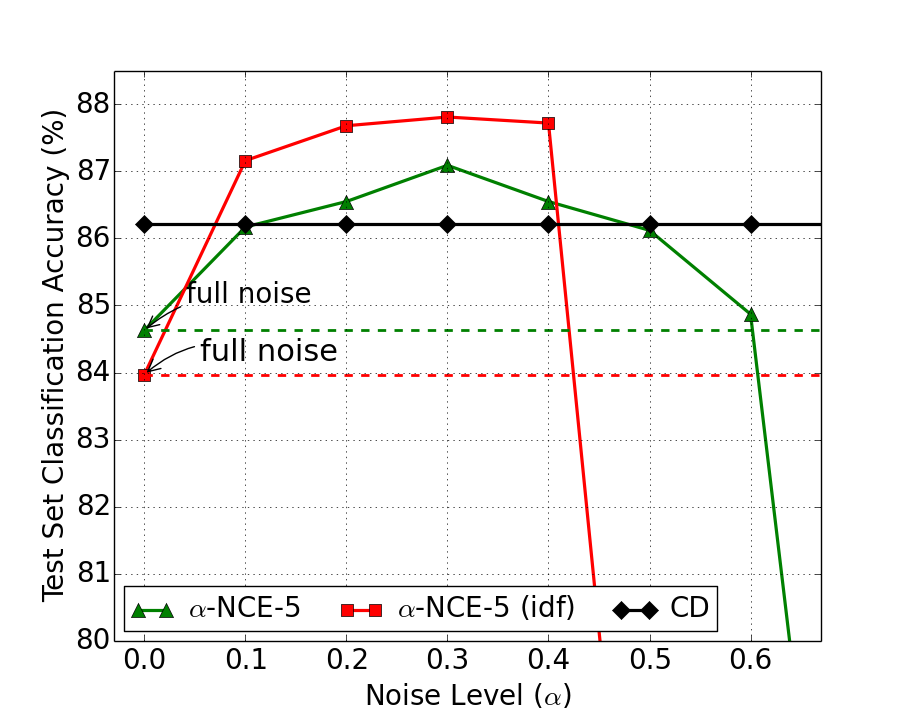}
}
\caption{Tracking the modelling performance with variant $\alpha$ using $\alpha$-NCE to learn RSMs. CD is also reported as the baseline. (a) (b) are performed on 20 Newsgroups, and (c) is performed on IMDB.}
\end{figure*}

For 20 Newsgroups, we trained RSMs on the training set, and reported the results on document retrieval and document classification. 
For retrieval, we treated the testing set as queries, and retrieved documents with the same labels in the training set by \textit{cosine-similarity}. 
Precision-recall (P-R) curves and mean average precision (MAP) are two metrics we used for evaluation. 
For classification, we trained a softmax regression on the training set, and checked the accuracy on the testing set.
We use this dataset to show the modelling ability of RSM with different estimators.

For IMDB, the whole training set is used for learning RSMs, and an L2-regularized logistic regression is trained on the labeled training set. The error rate of sentiment classification on the testing set is reported, compared with several BoW-based baselines. We use this dataset to show the general modelling ability of RSM compared with others. 

We trained both 
$\alpha$-NCE
and CD, and naturally NCE (without UCE) at a fixed vocabulary size (2000 for 20 Newsgroups, and 5000 for IMDB). Posteriors of the hidden units were used as topic features. For 
$\alpha$-NCE
, we fixed noise level at $0.5$ for 20 Newsgroups and $0.3$ for IMDB. In comparison, we trained CD from 1 up to 5 Gibbs steps.

Figure \ref{pr} and Table \ref{perplexity} show that a larger noise size in 
$\alpha$-NCE
achieves better modelling performance, and $\alpha$-NCE greatly outperforms CD on retrieval tasks especially around large recall values. The classification results of 
$\alpha$-NCE
is also comparable or slightly better than CD. 
Simultaneously, it is gratifying to find that the \textit{idf}-weighting inputs achieve the best results both in retrieval and classification tasks, as \textit{idf}-weighting is known to extract information better than word count.
In addition, naturally NCE performs poorly compared to others in Figure \ref{pr}, indicating variant document lengths actually bias the learning greatly. 
\begin{table}[htb]
\begin{small}
\begin{center}
\begin{tabular}{ccccc}
\hline 
\multirow{2}{*}{CD} &
\multicolumn{4}{c}{$\alpha$-NCE} \\
\cline{2-5}   
& k=1 & k=5 & k=25 & k=25 (idf)\\
\hline 
64.1\%  & 61.8\% & 63.6\% & \bf 64.8\% & \bf 65.6\% \\  
\hline
\end{tabular}
\end{center}
\caption{\label{perplexity} Comparison of classification accuracy on the 20 Newsgroups dataset using RSMs.}
\end{small}
\end{table}
\begin{table}[htb]
\begin{small}
\begin{center}
\begin{tabular}{l|c}
\hline
Models & Accuracy \\
\hline \hline
Bag of Words (BoW)  \cite{maas2010probabilistic} & 86.75\%   \\
LDA   \cite{maas2011learning} & 67.42\% \\
LSA   \cite{maas2011learning} & 83.96\% \\
\newcite{maas2011learning}'s ``full" model & 87.44\% \\
WRRBM \cite{dahl2012training} & 87.42\%   \\ 
\hline
RSM:CD & 86.22\% \\
RSM:$\alpha$-NCE-5 & \textbf{87.09\%} \\
RSM:$\alpha$-NCE-5 (idf) & \textbf{87.81\%} \\
\hline
\end{tabular}
\end{center}
\caption{\label{sent} The performance of sentiment classification accuracy on the IMDB dataset using RSMs compared to other BoW-based approaches.}
\end{small}
\end{table}

On the other hand, Table \ref{sent} shows the performance of RSM in sentiment classification, 
where model combinations reported in previous efforts are not considered. It is clear that 
$\alpha$-NCE learns RSM better than CD, and outperforms BoW and other BoW-based models\footnote{Accurately, WRRBM uses ``bag of \textit{n}-grams" assumption.} such as LDA. The \textit{idf}-weighting inputs also achieve the best performance.
Note that RSM is also based on BoW, indicating $\alpha$-NCE has arguably reached the limits of learning BoW-based models. In future work, RSM can be extended to more powerful undirected topic models, by considering more syntactic information such as word-order or dependency relationship in representation.
$\alpha$-NCE can be used to learn them efficiently and achieve better performance.

\subsection{Choice of Noise Level-$\bm{\alpha}$}
In order to decide the best noise level ($\alpha$) for PNS, we learned RSMs using 
$\alpha$-NCE
with different noise levels for both word count and \textit{idf}-weighting inputs on the two datasets.
Figure 3 shows that $\alpha$-NCE learning with partial noise ($\alpha>0$) outperforms full noise ($\alpha=0$) in most situations, and achieves better results than CD in retrieval and classification on both datasets.
However, 
learning tends to become extremely difficult if the noise becomes too close to the data, 
and this explains why the performance drops rapidly when $\alpha \rightarrow 1$.
Furthermore,
curves in Figure 3 also imply the choice of $\alpha$ might be problem-dependent, with larger sets like IMDB requiring relatively smaller $\alpha$. Nonetheless, a systematic strategy for choosing optimal $\alpha$ will be explored in future work. In practise, a range from $0.3 \sim 0.5$ is recommended.

\section{Conclusions}
We propose a novel approach $\alpha$-NCE for learning undirected topic models such as RSM efficiently, allowing large vocabulary sizes. It is new a estimator based on NCE, and adapted to documents with variant lengths and weighted inputs.
We learn RSMs with $\alpha$-NCE on two classic benchmarks, where it achieves both efficiency in learning and accuracy in retrieval and classification tasks.

\bibliography{acl2015}

\begin{thebibliography}{}

\bibitem[\protect\citename{Bengio \bgroup et al.\egroup
  }2003]{bengio2003neural}
Yoshua Bengio, R{\'e}jean Ducharme, Pascal Vincent, and Christian Janvin.
\newblock 2003.
\newblock A neural probabilistic language model.
\newblock {\em The Journal of Machine Learning Research}, 3:1137--1155.

\bibitem[\protect\citename{Bergstra \bgroup et al.\egroup
  }2010]{bergstra+al:2010-scipy}
James Bergstra, Olivier Breuleux, Fr{\'{e}}d{\'{e}}ric Bastien, Pascal Lamblin,
  Razvan Pascanu, Guillaume Desjardins, Joseph Turian, David Warde-Farley, and
  Yoshua Bengio.
\newblock 2010.
\newblock Theano: a {CPU} and {GPU} math expression compiler.
\newblock In {\em Proceedings of the Python for Scientific Computing Conference
  ({SciPy})}, June.
\newblock Oral Presentation.

\bibitem[\protect\citename{Blei \bgroup et al.\egroup }2003]{blei2003latent}
David~M Blei, Andrew~Y Ng, and Michael~I Jordan.
\newblock 2003.
\newblock Latent dirichlet allocation.
\newblock {\em the Journal of machine Learning research}, 3:993--1022.

\bibitem[\protect\citename{Dahl \bgroup et al.\egroup }2012]{dahl2012training}
George~E Dahl, Ryan~P Adams, and Hugo Larochelle.
\newblock 2012.
\newblock Training restricted boltzmann machines on word observations.
\newblock {\em arXiv preprint arXiv:1202.5695}.

\bibitem[\protect\citename{Dauphin and Bengio}2013]{dauphin2013stochastic}
Yann Dauphin and Yoshua Bengio.
\newblock 2013.
\newblock Stochastic ratio matching of rbms for sparse high-dimensional inputs.
\newblock In {\em Advances in Neural Information Processing Systems}, pages
  1340--1348.

\bibitem[\protect\citename{Gutmann and Hyv{\"a}rinen}2010]{gutmann2010noise}
Michael Gutmann and Aapo Hyv{\"a}rinen.
\newblock 2010.
\newblock Noise-contrastive estimation: A new estimation principle for
  unnormalized statistical models.
\newblock In {\em International Conference on Artificial Intelligence and
  Statistics}, pages 297--304.

\bibitem[\protect\citename{Hinton and Salakhutdinov}2009]{hinton2009replicated}
Geoffrey~E Hinton and Ruslan~R Salakhutdinov.
\newblock 2009.
\newblock Replicated softmax: an undirected topic model.
\newblock In {\em Advances in neural information processing systems}, pages
  1607--1614.

\bibitem[\protect\citename{Hinton}2002]{hinton2002training}
Geoffrey Hinton.
\newblock 2002.
\newblock Training products of experts by minimizing contrastive divergence.
\newblock {\em Neural computation}, 14(8):1771--1800.

\bibitem[\protect\citename{Hinton}2010]{hinton2010practical}
Geoffrey Hinton.
\newblock 2010.
\newblock A practical guide to training restricted boltzmann machines.
\newblock {\em Momentum}, 9(1):926.

\bibitem[\protect\citename{Hofmann}2000]{hofmann2000learning}
Thomas Hofmann.
\newblock 2000.
\newblock Learning the similarity of documents: An information-geometric
  approach to document retrieval and categorization.

\bibitem[\protect\citename{Hyv{\"a}rinen}2007]{hyvarinen2007some}
Aapo Hyv{\"a}rinen.
\newblock 2007.
\newblock Some extensions of score matching.
\newblock {\em Computational statistics \& data analysis}, 51(5):2499--2512.

\bibitem[\protect\citename{Kronmal and Peterson~Jr}1979]{kronmal1979alias}
Richard~A Kronmal and Arthur~V Peterson~Jr.
\newblock 1979.
\newblock On the alias method for generating random variables from a discrete
  distribution.
\newblock {\em The American Statistician}, 33(4):214--218.

\bibitem[\protect\citename{Maas and Ng}2010]{maas2010probabilistic}
Andrew~L Maas and Andrew~Y Ng.
\newblock 2010.
\newblock A probabilistic model for semantic word vectors.
\newblock In {\em NIPS Workshop on Deep Learning and Unsupervised Feature
  Learning}.

\bibitem[\protect\citename{Maas \bgroup et al.\egroup }2011]{maas2011learning}
Andrew~L Maas, Raymond~E Daly, Peter~T Pham, Dan Huang, Andrew~Y Ng, and
  Christopher Potts.
\newblock 2011.
\newblock Learning word vectors for sentiment analysis.
\newblock In {\em Proceedings of the 49th Annual Meeting of the Association for
  Computational Linguistics: Human Language Technologies-Volume 1}, pages
  142--150. Association for Computational Linguistics.

\bibitem[\protect\citename{Mikolov \bgroup et al.\egroup
  }2013a]{mikolov2013efficient}
Tomas Mikolov, Kai Chen, Greg Corrado, and Jeffrey Dean.
\newblock 2013a.
\newblock Efficient estimation of word representations in vector space.
\newblock {\em arXiv preprint arXiv:1301.3781}.

\bibitem[\protect\citename{Mikolov \bgroup et al.\egroup
  }2013b]{mikolov2013distributed}
Tomas Mikolov, Ilya Sutskever, Kai Chen, Greg~S Corrado, and Jeff Dean.
\newblock 2013b.
\newblock Distributed representations of words and phrases and their
  compositionality.
\newblock In {\em Advances in Neural Information Processing Systems}, pages
  3111--3119.

\bibitem[\protect\citename{Mnih and Hinton}2009]{mnih2009scalable}
Andriy Mnih and Geoffrey~E Hinton.
\newblock 2009.
\newblock A scalable hierarchical distributed language model.
\newblock In {\em Advances in neural information processing systems}, pages
  1081--1088.

\bibitem[\protect\citename{Mnih and Teh}2012]{mnih2012fast}
Andriy Mnih and Yee~Whye Teh.
\newblock 2012.
\newblock A fast and simple algorithm for training neural probabilistic
  language models.
\newblock {\em arXiv preprint arXiv:1206.6426}.

\bibitem[\protect\citename{Morin and Bengio}2005]{morin2005hierarchical}
Frederic Morin and Yoshua Bengio.
\newblock 2005.
\newblock Hierarchical probabilistic neural network language model.
\newblock In {\em Proceedings of the international workshop on artificial
  intelligence and statistics}, pages 246--252. Citeseer.

\bibitem[\protect\citename{Salakhutdinov and
  Hinton}2009]{salakhutdinov2009semantic}
Ruslan Salakhutdinov and Geoffrey Hinton.
\newblock 2009.
\newblock Semantic hashing.
\newblock {\em International Journal of Approximate Reasoning}, 50(7):969--978.

\bibitem[\protect\citename{Salton and McGill}1983]{salton1983introduction}
Gerard Salton and Michael~J McGill.
\newblock 1983.
\newblock Introduction to modern information retrieval.

\bibitem[\protect\citename{Srivastava \bgroup et al.\egroup
  }2013]{srivastava2013modeling}
Nitish Srivastava, Ruslan~R Salakhutdinov, and Geoffrey~E Hinton.
\newblock 2013.
\newblock Modeling documents with deep boltzmann machines.
\newblock {\em arXiv preprint arXiv:1309.6865}.

\bibitem[\protect\citename{Tieleman}2008]{tieleman2008training}
Tijmen Tieleman.
\newblock 2008.
\newblock Training restricted boltzmann machines using approximations to the
  likelihood gradient.
\newblock In {\em Proceedings of the 25th international conference on Machine
  learning}, pages 1064--1071. ACM.

\end{thebibliography}
\bibliographystyle{acl}

\end{document}